\documentclass[journal]{IEEEtran}

\usepackage{lipsum}
\usepackage{array}
\usepackage{times}
\usepackage{epsfig}
\usepackage{graphicx}
\usepackage{float}
\usepackage{wrapfig}
\usepackage{amsmath,amssymb,amsthm}
\usepackage{algorithm,algorithmicx,algpseudocode}
\usepackage{bm,xspace}
\usepackage{comment}
\usepackage{multirow}
\usepackage{balance}
\usepackage{url}
\usepackage{booktabs}
\usepackage{etoolbox,siunitx}
\usepackage{calc}
\usepackage{pifont,hologo}
\usepackage{color}
\usepackage{adjustbox}
\usepackage{enumitem}
\usepackage{bbding}  %
\usepackage[normalem]{ulem}  %
\usepackage{contour}
\usepackage[table]{xcolor}
\usepackage{soul}%
\usepackage{tocloft} %
\usepackage{etoc} %
\PassOptionsToPackage{dvipsnames}{xcolor}
\usepackage[breaklinks,colorlinks,linkcolor=link,citecolor=cite]{hyperref} % pagebackref
\PassOptionsToPackage{square,numbers,comma,sort}{natbib}
\usepackage{natbib}
\usepackage[caption=false,font=normalsize,labelfont=sf,textfont=sf]{subfig}
\usepackage{mathtools}
\usepackage{orcidlink}

\definecolor{blue}{HTML}{004bb3}
\definecolor{red}{HTML}{cc1100}
\definecolor{orange}{HTML}{cc7700}
\definecolor{gray}{HTML}{efefef}
\definecolor{darkgreen}{HTML}{228B22}
\definecolor{darkgray}{HTML}{757575}

\definecolor{cite}{HTML}{3270b5}
\definecolor{link}{HTML}{b53532}
\definecolor{link}{HTML}{cc1100}
\definecolor{scratch}{HTML}{001219}
\definecolor{pretrain}{HTML}{0A9396}

\contourlength{0.8pt}

\newcommand{\figref}[1]{Fig.~\ref{#1}}
\newcommand{\tabref}[1]{Tab.~\ref{#1}}
\newcommand{\secref}[1]{Sec.~\ref{#1}}
\renewcommand{\eqref}[1]{Eq.~\ref{#1}}

\newcolumntype{x}[1]{>{\centering\arraybackslash}p{#1}}
\newcolumntype{y}[1]{>{\raggedright\arraybackslash}p{#1}}
\newcolumntype{z}[1]{>{\raggedleft\arraybackslash}p{#1}}
\newcommand{\tablestyle}[2]{\setlength{\tabcolsep}{#1}\renewcommand{\arraystretch}{#2}\centering\footnotesize}

\DeclareMathSymbol{@}{\mathord}{letters}{"3B}

\newcommand{\YesV}{\ding{51}}%
\newcommand{\NoX}{\ding{55}}%

\makeatletter
\DeclareRobustCommand\onedot{\futurelet\@let@token\@onedot}
\def\@onedot{\ifx\@let@token.\else.\null\fi\xspace}
\def\eg{\emph{e.g}\onedot} 
 
\def\cf{\emph{cf}\onedot}

\newcommand*{\Rom}[1]{\expandafter\@slowromancap\romannumeral #1@}
\newcommand*{\rom}[1]{\expandafter\romannumeral #1}

\def\1{\bm{1}}

\def\rvf{{\mathbf{f}}}

\def\rvs{{\mathbf{s}}}

\def\rmE{{\mathbf{E}}}
\def\rmF{{\mathbf{F}}}

\def\rmQ{{\mathbf{Q}}}

\def\rmS{{\mathbf{S}}}

\def\rmW{{\mathbf{W}}}

\def\vj{{\bm{j}}}
\def\vk{{\bm{k}}}

\def\vv{{\bm{v}}}

\def\mQ{{\bm{Q}}}

\DeclareMathAlphabet{\mathsfit}{\encodingdefault}{\sfdefault}{m}{sl}
\SetMathAlphabet{\mathsfit}{bold}{\encodingdefault}{\sfdefault}{bx}{n}

\def\gA{{\mathcal{A}}}

\def\gC{{\mathcal{C}}}

\def\gF{{\mathcal{F}}}
\def\gG{{\mathcal{G}}}

\def\gI{{\mathcal{I}}}
\def\gJ{{\mathcal{J}}}

\def\gL{{\mathcal{L}}}
\def\gM{{\mathcal{M}}}
\def\gN{{\mathcal{N}}}
\def\gO{{\mathcal{O}}}
\def\gP{{\mathcal{P}}}
\def\gQ{{\mathcal{Q}}}

\def\gS{{\mathcal{S}}}

\def\gU{{\mathcal{U}}}
\def\gV{{\mathcal{V}}}

\newcommand{\R}{\mathbb{R}}

\let\originalleft\left
\let\originalright\right
\renewcommand{\left}{\mathopen{}\mathclose\bgroup\originalleft}
\renewcommand{\right}{\aftergroup\egroup\originalright}

\newcommand{\ours}{SGIFormer\xspace}
\newcommand{\scannet}{ScanNet V2\xspace}
\newcommand{\scannetp}{ScanNet200\xspace}

\newcommand{\scannetpp}{ScanNet++\xspace}
\newcommand{\smq}{SMQ\xspace}
\newcommand{\git}{GIT\xspace}
\newcommand{\spformer}{SPFormer~\cite{sun2023spformer}\xspace}
\newcommand{\sph}{Spherical Mask~\cite{shin2023spherical}\xspace}

\begin{document}

\title{\ours: Semantic-guided and Geometric-enhanced Interleaving Transformer for 3D Instance Segmentation}

\author{Lei Yao~\orcidlink{0009-0007-0304-3056},
        Yi Wang~\orcidlink{0000-0001-8659-4724},~\IEEEmembership{Member,~IEEE,}
        Moyun Liu~\orcidlink{0000-0002-4530-2606},
        and~Lap-Pui Chau~\orcidlink{0000-0003-4932-0593},~\IEEEmembership{Fellow,~IEEE}% <-this % stops a space
        
\thanks{Manuscript received xx, xx; revised xx, xx. \emph{(Corresponding author: Lap-Pui Chau)}}

\thanks{Lei Yao, Yi Wang, and Lap-Pui Chau are with the Department of Electrical and Electronic Engineering, The Hong Kong Polytechnic University, Hong Kong SAR (e-mail: rayyoh.yao@connect.polyu.hk; yi-eie.wang@polyu.edu.hk;
lap-pui.chau@polyu.edu.hk).}% <-this % stops a space

\thanks{Moyun Liu is with the School of Mechanical Science and Engineering, Huazhong University of Science and Technology, Wuhan 430074, China (e-mail: lmomoy@hust.edu.cn).}
}

\markboth{Journal of \LaTeX\ Class Files,~Vol.~14, No.~8, August~2021}%
{Shell \MakeLowercase{\textit{et al.}}: Bare Demo of IEEEtran.cls for IEEE Journals}

\maketitle

\begin{abstract}
    % Inspired by recent advances in 2D segmentation, 
In recent years, transformer-based models have exhibited considerable potential in point cloud instance segmentation. Despite the promising performance achieved by existing methods, they encounter challenges such as instance query initialization problems and excessive reliance on stacked layers, rendering them incompatible with large-scale 3D scenes. This paper introduces a novel method, named \ours, for 3D instance segmentation, which is composed of the Semantic-guided Mix Query (\smq) initialization and the Geometric-enhanced Interleaving Transformer (\git) decoder. Specifically, the principle of our \smq initialization scheme is to leverage the predicted voxel-wise semantic information to implicitly generate the scene-aware query, yielding adequate scene prior and compensating for the learnable query set. %detail retention. 
%Subsequently, by incorporating another learnable query set, 
Subsequently, we feed the formed overall query into our \git decoder to alternately refine instance query and global scene features for further capturing fine-grained information and reducing complex design intricacies simultaneously. To emphasize geometric property, we consider bias estimation as an auxiliary task and progressively integrate shifted point coordinates embedding to reinforce instance localization. \ours attains state-of-the-art performance on \scannet, \scannetp datasets, and the challenging high-fidelity \scannetpp benchmark, striking a balance between accuracy and efficiency. The code, weights, and demo videos are publicly available at \href{https://rayyoh.github.io/sgiformer/}{https://rayyoh.github.io/sgiformer}.
\end{abstract}

\begin{IEEEkeywords}
    Point Clouds, 3D Instance Segmentation, Transformer, Semantic Features
\end{IEEEkeywords}

\section{Introduction}\label{sec:introduction}

\IEEEPARstart{P}{oint} cloud instance segmentation serves as a fundamental task for 3D scene understanding across various applications such as embodied AI~\cite{delitzas2024scenefun3d, qi2023instance}, autonomous driving~\cite{zhou2020joint, liu2024menet}, and metaverse~\cite{wirth2019pointatme}. The primary objective of this task is to identify each instance using binary masks and assign corresponding semantic categories within a scanned scene. However, due to the unordered nature of points and the sophisticated layout of scenes, accurately segmenting objects with proximity and varying sizes remains challenging in 3D point cloud instance segmentation.

\begin{figure}[tb] 
    \centering
    \begin{minipage}[t]{0.48\textwidth}
        \centering
        \includegraphics[width=0.9\textwidth]{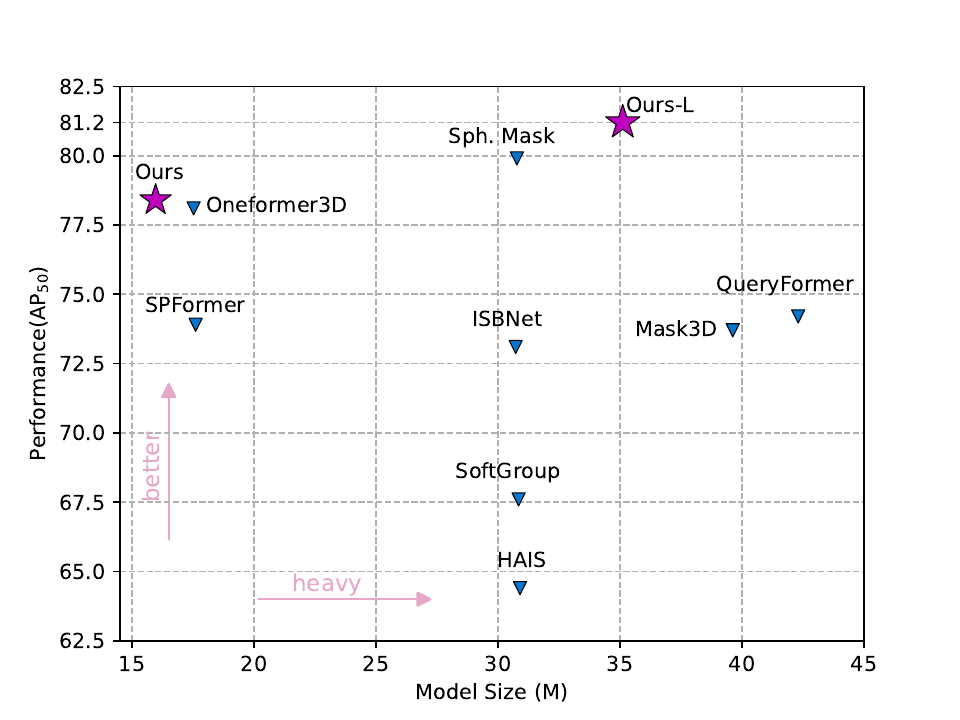} \\
        \scriptsize (a) Model size vs. instance segmentation performance AP$_{50}$.
    \end{minipage}\hfill
    \begin{minipage}[t]{0.48\textwidth}
        \centering
        \includegraphics[width=0.9\textwidth]{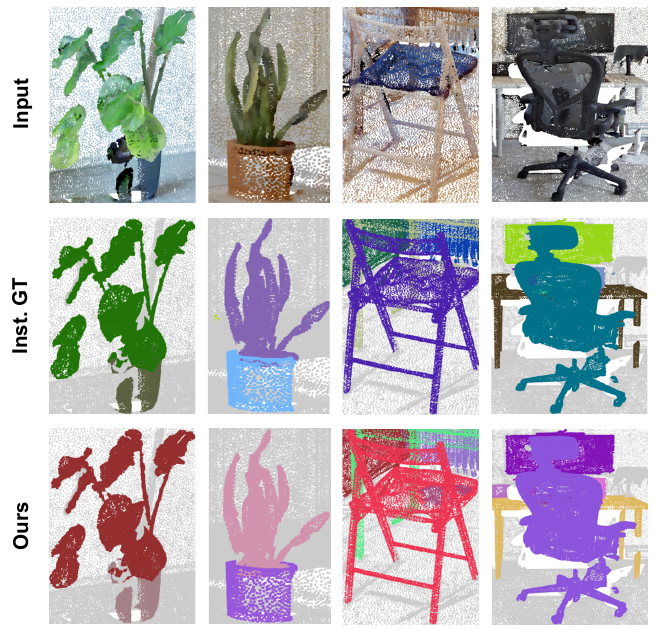} \\
        \scriptsize (b) Examples of fine-grained segmentation on the \scannetpp validation set.
    \end{minipage}
    
    \caption{\textbf{Performance evaluation of our proposed \ours.} (a) We present the performance comparison of various methods based on AP$_{50}$ and model size on \scannet~\cite{dai2017scannet} validation split. \ours outperforms previous methods, and even the smaller version achieves competitive results. (b) We showcase the fine-grained segmentation results of \ours on \scannetpp~\cite{yeshwanth2023scannet++} validation set, demonstrating its ability to segment small objects within large-scale scenes accurately.} 
    \label{fig:teaser}
\end{figure}

Current approaches for point cloud instance segmentation can be categorized into three distinct pipelines: proposal-based~\cite{hou20193dsis, yi2019gspn, yang2019learning, kolodiazhnyi2023td3d, shin2023spherical}, grouping-based~\cite{jiang2020pointgroup, chen2021hais, vu2022softgroup, liang2021sstnet, han2020occuseg, hui2022learning, zhao2022jsnet++}, and transformer-based~\cite{sun2023spformer, schult2023mask3d, lu2023query, lai2023mask,kolodiazhnyi2023of3d, al20233d, zhao2023eipformer}. Proposal-based methods follow a top-down strategy by generating a series of preliminary proposals and refining them to obtain accurate instance masks. Grouping-based methods directly aggregate points into instance clusters according to point-wise semantic information and positional relationships in a bottom-up manner. However, both proposal-based and grouping-based architectures share a common limitation: reliance on accurate intermediate results. For example, incorrect object bounding boxes or false class labels~\cite{jiang2020pointgroup} may lead to error accumulation during subsequent processes, resulting in suboptimal performance. In contrast, transformer-based methods operate in a fully end-to-end style, leveraging the attention mechanism~\cite{vaswani2017attention} to capture global context information of the scene efficiently. This approach typically utilizes a 3D backbone to extract global scene features from the input point cloud and feed them into a stacked transformer decoder with a fixed number of instance queries to refine them iteratively. The method regards each query as a potential object instance and simultaneously predicts the corresponding mask and category.

While transformer-based methods have shown great promise in 3D instance segmentation and achieved superior performance with compact pipelines, existing algorithms still have inherent limitations. Considering that the quality of the query affects both the final performance and the convergence speed of the model, we suggest that \textit{query initialization} is the first crucial component that requires improvement. As pioneers in this field, \spformer and Mask3D~\cite{schult2023mask3d} have attempted different strategies for query initialization. The former employs randomly initialized parametric queries that are learnable, while the latter samples queries from the raw input, which are non-parametric, and it observes performance improvement similar to~\cite{misra2021end}. However, current sampling methods such as farthest point sampling (FPS)~\cite{qi2017pointnet++} cannot guarantee high-quality queries, as the sampled points may overlook small instances or even be located in non-informative background regions~\cite{yin2023dcnet}. Additionally, the uncertainty in FPS might result in multiple queries covering the same object, further decreasing the quality of segmentation tasks. Another limitation lies in the \textit{deficiency of geometric information and fine-grained details} during the query refinement stage. The decoder was initially designed to update instance queries by aggregating the scene features. Nevertheless, due to the quadratic complexity of the attention mechanism~\cite{vaswani2017attention}, the features are often pooled from point-level embedding~\cite{sun2023spformer, kolodiazhnyi2023of3d}. Although this reduces the computational cost, it neglects the fine-grained details of the original scene, which has not been adequately addressed in previous works.

Motivated by the abovementioned analysis, this paper presents a \textbf{S}emantic-guided and \textbf{G}eometric-enhanced \textbf{I}nterleaving Trans\textbf{Former} (\ours) for point cloud instance segmentation, which aims to overcome the limitations of existing transformer-based methods. Concretely, we propose a Semantic-guided Mix Query (\smq) initialization scheme by incorporating voxel-wise semantic prediction as guidance to filter out weak semantic regions and generate scene-aware queries from remaining voxels, maintaining scene prior and local details. Additionally, we combine another set of learnable queries to form the overall query set, enhancing the diversity and adaptability. To emphasize geometric information, we introduce a Geometric-enhanced Interleaving Transformer (\git) decoder to gradually participate in point coordinates embedding into global scene features and update them as well as queries alternately to avoid the loss of details. With diverse queries and the interleaving mechanism, our \ours can benefit from the local semantic information and efficiently reduce the reliance on heavy stacked layers, improving both efficiency and accuracy. The main contributions of this work can be summarized as follows:
\begin{itemize}
    \item We propose a novel query initialization scheme to obtain an instance query set with semantic guidance. This scheme effectively integrates scene prior and preserves local information, improving the quality and adaptability of queries.
    \item We introduce an interleaving transformer decoder progressively incorporating geometric information to refine instance queries and global scene features alternately, reducing the reliance on heavy stacked layers and enhancing the preservation of fine-grained details.
    \item Comprehensive experiments are conducted on various datasets, including \scannet, \scannetp, and \scannetpp, to evaluate the proposed \ours. The experimental results demonstrate the superiority of our method over existing state-of-the-art methods, establishing its effectiveness in 3D instance segmentation.
\end{itemize}

\section{Related work}\label{sec:related}

\subsection{3D Instance Segmentation Methods}
3D instance segmentation methods can be broadly categorized into proposal-based, grouping-based, and transformer-based methods.

% Top-down proposal-based methods coarse-to-fine
Proposal-based methods~\cite{hou20193dsis, yi2019gspn, yang2019learning, kolodiazhnyi2023td3d, shin2023spherical} generate object proposals at the first phase following a refinement stage to obtain instance masks in a top-down manner. Taking RGB-D scans as input, 3D-SIS~\cite{hou20193dsis} used predicted 3D bounding boxes to acquire associated fine masks. 3D-BoNet~\cite{yang2019learning} conducted mask prediction by concatenating intermediate results and per-point features after getting instance bounding boxes using the Hungarian matching~\cite{kuhn1955hungarian} algorithm.  Following a similar style, TD3D~\cite{kolodiazhnyi2023td3d} proposed a data-driven, fully convolutional network without relying on prior knowledge or handcrafted parameters. Differently, GSPN~\cite{yi2019gspn} adopted an analysis-by-synthesis strategy by reconstructing shapes to get proposals. Due to the inherent dependence on object proposals, false negative error might accumulate from inaccurate bounding box prediction in the abovementioned methods. Therefore, Spherical Mask~\cite{shin2023spherical} represented a 3D polygon using spherical coordinates for coarse instance detection and incorporated a point migration module to alleviate error propagation. However, the complicated multi-stage nature and post-processing steps lead to significant latency overhead. Instead, our \ours benefits from the advantages of end-to-end set prediction and avoids redundant computation.

% Bottom-up Grouping-based Methods
Grouping-based methods %%in point cloud instance segmentation aim to %%
cluster points or superpoints~\cite{felzenszwalb2004efficient, landrieu2018superpoint} into object instances based on corresponding semantic categories~\cite{jiang2020pointgroup, chen2021hais, vu2022softgroup, liang2021sstnet}, geometric offsets~\cite{jiang2020pointgroup, chen2021hais, vu2022softgroup} or feature similarity~\cite{han2020occuseg, hui2022learning}. With predicted points-wise class labels, PointGroup~\cite{jiang2020pointgroup} considered both shifted coordinates and original ones simultaneously to get groups followed by Non-Maximum Suppression (NMS) for final instance masks. %%, and 3DT-Seg~\cite{wang20233d} improved the dual aggregation strategy to be adaptive%% 
To further improve clustering accuracy, HAIS~\cite{chen2021hais} proposed hierarchical aggregation, which gathers points into a series of sets and obtains complete instances by set aggregation. SoftGroup~\cite{vu2022softgroup} introduced a soft grouping mechanism that allows each point to be assigned multiple categories, reducing errors from semantic prediction. In~\cite{zhao2023pbnet}, a binary clustering strategy is proposed as an alternative to traditional methods. %%Considering the significance of geometric prior,%%
SSTNet~\cite{liang2021sstnet} and GraphCut~\cite{hui2022learning} utilized superpoints as a mid-level representation and directly merged them into instances using semantic tree structure or graph neural networks. By eliminating the need for fancy offline clustering operation, our \ours offers a more streamlined architecture %%and achieves end-to-end instance segmentation.

%% The attention mechanism~\cite{vaswani2017attention} introduced in Transformer has proven to be highly effective and versatile in various domains, including natural language processing (NLP) and computer vision (CV)%%. 
Building upon the flexibility and superiority of Transformer, DETR~\cite{carion2020detr} introduced a compact set prediction pipeline for object detection, which was later extended to dense segmentation tasks~\cite{cheng2021maskformer, cheng2022mask2former}. Inspired by DETR~\cite{carion2020detr}, a series of works have adapted the paradigm to 3D domain for object detection~\cite{wang2022cagroup3d, liu2021groupfree} and instance segmentation~\cite{sun2023spformer, schult2023mask3d, lu2023query, lai2023mask, kolodiazhnyi2023of3d, al20233d}. \spformer and Mask3D~\cite{schult2023mask3d} are pioneers in achieving end-to-end 3D instance segmentation by iteratively refining a fixed number of instance queries. Additionally, 3IS-ESS~\cite{al20233d} pointed out that the current pipeline lacks information exchange between the backbone and query refinement decoder. To moderate this, they proposed using voxel-wise semantic label prediction and raw coordinate regression as auxiliary tasks to enhance semantic and spatial understanding. Nevertheless, the intermediate results in their method are underutilized. In contrast, we propose using sufficient semantic cues to guide scene-aware query initialization.

\subsection{Query Initialization and Refinement}
Although the end-to-end transformer-based architecture is elegant, one crucial challenge is efficiently initializing queries to accelerate the training process further and boost the final performance. Similar to DETR~\cite{carion2020detr} and Mask2Former~\cite{cheng2022mask2former}, parametric learnable queries are used in \spformer and MAFT~\cite{lai2023mask}, while other approaches like Mask3D~\cite{schult2023mask3d} demonstrated that sampling points from raw input scene according to their coordinates to initialize queries can lead to improved performance, as observed by ~\cite{misra2021end} in 2D domain. Following this configuration, QueryFormer~\cite{lu2023query} proposed a tailored query aggregation module to reduce duplicate queries belonging to the same instance, aiming to increase coverage, and LCPFormer~\cite{huang2023lcpformer} utilized local context propagation strategy. However, these modules introduced additional computational expenditure with limited benefits. OneFormer3D~\cite{kolodiazhnyi2023of3d} randomly selected features from available over-segments of \scannet~\cite{dai2017scannet} as queries for training, but for better performance, all superpoints were used during inference. Even though this approach achieved one model for three different segmentation tasks on a specific dataset, it does not readily generalize to other datasets, as it showed a significant performance decline on S3DIS~\cite{armeni2016s3dis}. Differently, we introduce a novel mix query initialization strategy that combines scene-aware query and learnable query to achieve better generalization and performance. 

For query refinement, existing methods~\cite{sun2023spformer, lai2023mask, schult2023mask3d, kolodiazhnyi2023of3d} typically adopt standard transformer decoders to progressively attend to features from superpoints or voxels, guided by masked attention~\cite{cheng2022mask2former}, to enhance the stability of the training process. These approaches rely on heavily stacked transformer layers and do not consider the built-in advantages of geometric property for point cloud data in the refinement process. However, our \ours participates in shifted point coordinates embedding to reinforce the global features for better instance localization progressively and adopts a novel interleaving update mechanism to alternately refine instance query and global scene features for more effective information exchange.

\section{Methodology}\label{sec:method}

\begin{figure*}[!htbp]
    \centering
    \includegraphics[width=\textwidth]{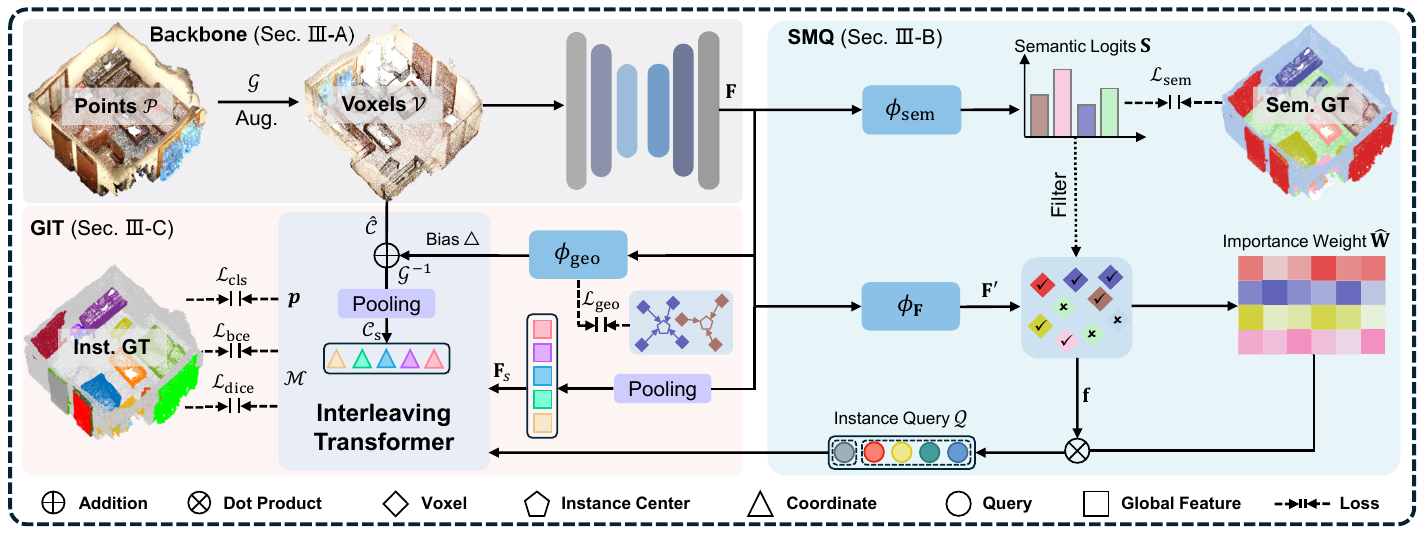}
    \caption{\textbf{Overall architecture of \ours.} The method comprises three main components: a symmetrical U-Net backbone, a Semantic-guided Mix Query (\smq) initialization scheme, and a Geometric-enhanced Interleaving Transformer (\git) decoder. Aug. in this figure denotes data augmentation. The 3D backbone extracts voxel-wise global features $\rmF$ from the input point cloud $\gP$ (\secref{subsec:backbone}). \smq constructs instance queries $\gQ$ with semantic guidance (\secref{subsec:query}). \git alternately refines the queries and scene features to enhance geometric information and capture fine-grained details. The final instance masks $\gM$ and categories $\boldsymbol{p}$ are predicted by the decoder (\secref{subsec:decoder}).}
    \label{fig:overall}
\end{figure*}

In this section, we provide an overview of our pipeline and discuss the details of our methodology. We begin by describing the input data and backbone for feature extraction in \secref{subsec:backbone}. We then elaborate on our novel semantic-guided mix query initialization scheme and geometric-enhanced interleaving transformer decoder in \secref{subsec:query} and \secref{subsec:decoder}, respectively. Finally, we discuss the loss function used in our method in \secref{subsec:loss}.

\subsection{Backbone}\label{subsec:backbone}

Given a series of point cloud coordinates $\gC \in \R^{n \times 3}$ and corresponding $r,g,b$ colors $\gF \in \R^{n \times 3}$, where $n$ is the number of points. Our method takes a scene point set $\gP = \{\gC, \gF\}$ as input as illustrated in \figref{fig:overall}. Before feeding them into the 3D backbone, we perform quantization by dividing the input points into $m$ voxels $\gV$ by a grid sampling pattern $\gG: \gP \mapsto \gV=\{\hat{\gC}, \hat{\gF}\}$. Note that $\hat{\gC} \in \R^{m\times3}$ and $\hat{\gF} \in \R^{m\times3}$ are initialized by averaging the coordinates $\gC$ and colors $\gF$ of points within each voxel. Then the Submanifold Sparse Convolution~\cite{graham2017submanifold} is adopted to implement our symmetrical U-Net backbone to extract voxel-wise global features $\rmF \in \R^{m \times d_o}$ similar to~\cite{sun2023spformer, shin2023spherical}, where $d_o$ is the channel dimension of the features. 

\subsection{Semantic-guided Mix Query Initialization}\label{subsec:query}

As a pivotal component of transformer-based decoders, query initialization has been explored in various 3D tasks~\cite{liu2021groupfree, kolodiazhnyi2023of3d, wang2024uni3detr}. Generally, the query can be divided into parametric and non-parametric. The former initializes queries with learnable parameters, while the latter employs sampled point features as queries. FPS~\cite{qi2017pointnet++} is a typical selection mechanism in the point cloud domain. Parametric queries are flexible but suffer from slow convergence, while non-parametric queries are efficient but lack specific guidance and may include background points, leading to suboptimal results~\cite{lu2023query}. To tackle these problems, in this work, we propose a novel Semantic-guided Mix Query (\smq) initialization scheme that implicitly obtains scene-aware queries, combining with learnable queries as the whole query set.

To achieve semantic guidance, we construct a branch that predicts voxel-wise class labels facilitating semantic prior formulated as $\rmS = \phi_{\texttt{sem}}(\rmF) \in \R^{m \times (c+1)}$, where $\rmS = \{\rvs_{i}\}_{i=1}^{m}$ and $c$ represents the number of instance categories. Here, an extra label $\varnothing$ indicates the background. With available voxel-level semantic information, we train this branch using cross-entropy loss:
\begin{equation}
    \gL_{\texttt{sem}} = -\frac{1}{|\gV|} \sum_{v_i \in \gV} \sum_{j \in \gJ} \mathbf{1}_{\{y_{v_i}=j\}} \log(s_{i,j}),
    \label{eq:loss_sem}
\end{equation}
where $s_{i,j}$ is the probability of voxel $v_i$ belonging to class $j$, $y_{v_i}$ is the ground truth label, $\gJ=\{1, \ldots, c+1\}$ is the set of class labels and $\mathbf{1}_{\{\cdot\}}$ denotes the indicator function.

To yield the scene-aware queries $\rmQ^{s} = \{\mQ^{s}_{1}, \ldots, \mQ^{s}_{q_s}\} \in \R^{q_s \times d}$ from voxel features $\rmF$, we propose an implicit query initialization algorithm with semantic prediction results as guidance outlined in Alg.~\ref{alg:query}. Initially, the method calculates semantic score $\rmS' \in \R^{m \times c}$ for each voxel, excluding the background, and selects top-k favorable voxels with high semantics. Rather than straightforwardly applying a fixed threshold, we dynamically adjust the number of selected voxels $\alpha m$ using a ratio $\alpha$ based on the scale $m$ of the scene. This adaptive approach enables us to efficiently filter out background noise and concentrate on foreground voxels $\rvf = \rmF'[idx] \in \R^{\alpha m \times d}$ with a higher probability of being relevant instances of interest. After the selection stage, the algorithm generates a sparse set of query weights $\rmW = \left[w_{u,v}\right]^{q_s, \alpha m}_{u=1,v=1}$ by:
\begin{equation}
    \rmW = \psi\left(\rvf\right)^\top,
    \label{eq:weight}
\end{equation}
where $\psi(\cdot): \R^{\alpha m \times d} \mapsto \R^{\alpha m \times q_s}$ is a voxel-wise feature transformation implemented by MLP: $\psi(\rvf)=\texttt{ReLU}(\boldsymbol{w}^\top_{\psi} \rvf+\boldsymbol{b}_{\psi})$. We then normalize the weight by \textit{softmax} function:
\begin{equation}
    \hat{w}_{u,v} = \frac{\exp(w_{u,v})}{\sum_{j=1}^{\alpha m} \exp(w_{u,j})},
    \label{eq:softmax}
\end{equation}
$\hat{w}_{u,v}$ represents the importance weight of the $v$-th selected voxel $\rvf_{v} \subseteq \rvf$ for the $u$-th query. For simplicity, we denote the above operation as $\mathbf{\gI}$. Finally, the initialized scene-aware query $\mathbf{Q}^{p}$ is formulated by summing the chosen voxel features with the normalized weights:
\begin{equation}
    \mQ^{s}_{u} = \sum_{v=1}^{\alpha m} \hat{w}_{u,v} \cdot \rvf_{v}.
    \label{eq:query}
\end{equation}
With the overall process, we can implicitly cluster the representative voxels into a set of scene-aware instance queries containing fine-grained foreground features. At the same time, our design can be more robust to noise in semantic prediction, avoiding potential error propagation indicated in~\cite{vu2022softgroup}.

\begin{algorithm}
    \caption{Implicit Scene-aware Query Initialization}
    \begin{algorithmic}[1]
        \Require voxel-wise features $\rmF$, semantic logits $\rmS$, ratio $\alpha$, \Statex \quad \quad { } number of scene-aware query $q_s$
        \Ensure initial scene-aware query $\rmQ^{s}$
        \State \textcolor{darkgray}{\text{/* get semantic score without background */}}
        \State $\rmS'\in \R^{m \times c} \gets \mathbf{softmax}(\rmS)[:, :-1]$
        \State $\vj \gets \mathbf{argmax}(\rmS', \mathbf{dim=}-1)$
        \State \textcolor{darkgray}{\text{/* filter out disruptive voxels */}}
        \State $score, idx \gets \mathbf{topk}(\rmS'[:, \vj], \alpha m)$
        \State \textcolor{darkgray}{\text{/* project original voxel features */}}
        \State $\rmF' \in \R^{m \times d} \gets \mathbf{Linear}(\rmF)$
        \State $\rvf \in \R^{\alpha m \times d}  \gets \rmF'[idx]$
        % \State \textcolor{darkgray}{\text{/* project selected voxel features to $\mathbf{\gI} \in \R^{\alpha m \times q_s}$ */}}
        % \State $\mathbf{\gI} \gets \mathbf{Linear_{\mathbf{\gI}}}(\mathbf{F'}[idx])$
        % \State \textcolor{darkgray}{\text{/* $\hat{\rmW}$ is the normalized weight of selected voxels */}}
        % \State $\hat{\rmW} \gets \mathbf{\gI}(\rvf)$
        \State \textcolor{darkgray}{\text{/* get initialized scene-aware query */}}
        \State $\rmQ^{s} \in \R^{q_s \times d} \gets \mathbf{\gI}(\rvf) \odot \rvf$
        \State \textbf{Return}: $\rmQ^{s}$
    \end{algorithmic}
    \label{alg:query}
\end{algorithm}

The final query set $\gQ = [\rmQ^{s}, \rmQ^{l}] \in \R^{q \times d}$ for transformer decoder is achieved by combining the scene-aware query $\rmQ^{s}$ with another set of randomly initialized learnable query $\rmQ^{l} \in \R^{q_l \times d}$, where $\left[\cdot, \cdot\right]$ denotes concatenation operation and $q = q_s + q_l$ means the total query number. The inclusion of parametric queries aims to capture missing local information and adapt the model to different scenes. We argue that our query initialization scheme benefits from both semantic cues and learnable queries. The guided queries can provide semantic prior through fine-grained voxel features and yield accelerating convergence, while additional parametric queries can improve the flexibility of the model. Furthermore, with the assistance of our implicitly generated queries, inter-query interaction during instance self-attention can extract more informative global contexts, reducing the dependence on heavily stacked layers. The mix query initialization module can achieve a balance between diverse queries and further promote subsequent transformer decoder.

\subsection{Geometric-enhanced Interleaving Transformer Decoder}\label{subsec:decoder}

Generally, in DETR-based 3D instance segmentation methods~\cite{sun2023spformer, kolodiazhnyi2023of3d, schult2023mask3d}, transformer decoder is utilized to refine instance queries with context information from the global scene features by stacking multiple layers. However, the vanilla decoder, whether for parametric query in~\cite{sun2023spformer, lai2023mask} or non-parametric query in~\cite{kolodiazhnyi2023of3d}, ignores the geometric property~\cite{lin2023dbganet} of the input point clouds during query refinement. These methods directly update the query by attending to features of coarse-grained superpoints, which leads to a lack of inter-superpoint communication and the potential loss of fine-grained details in the input scene. Recognizing this limitation, we propose an interleaving transformer decoder capable of capturing instance geometric and detailed information more efficiently.

We consider the coordinate of voxels $\hat{\gC}$ as the key component for geometry. Nevertheless, substantial variations in the scene range cause unstable training of raw voxel coordinates regression as in~\cite{al20233d}. Instead, we estimate the bias vectors $\mathbf{\Delta} = \phi_{\texttt{geo}}(\rmF) \in \R^{m \times 3}$ of each voxel relative to the instance geometric center it belongs to. Since ground-truth instance centers are available, we apply $\ell_1$ loss to supervise the geometric branch:
\begin{equation}
    \gL_{\texttt{geo}} = \frac{1}{\sum_{v_i \in \gV} \mathbf{1}_{\{v_i\}}} \sum_{v_i \in \gV} \mathbf{1}_{\{v_i\}}\|\mathbf{\Delta}_i - (\gO^{*}_i - \hat{\gC}_i)\|_1,
    \label{eq:loss_geo}
\end{equation}
where $\gO^{*}_i$ is the ground truth geometric center of the instance to which voxel $v_i$ belongs, $\mathbf{1}_{\{v_i\}}$ is the indicator function to determine whether a voxel belongs to an instance. Then we shift the learned bias to raw coordinates $\hat{\gC}$ to obtain refined coordinates, given by:
\begin{equation}
    \hat{\gC}_{\texttt{ref}} = \hat{\gC} + \mathbf{\Delta}.
    \label{eq:voxel_bias}
\end{equation}

\begin{figure}[htbp] 
    \centering
    \includegraphics[width=0.48\textwidth]{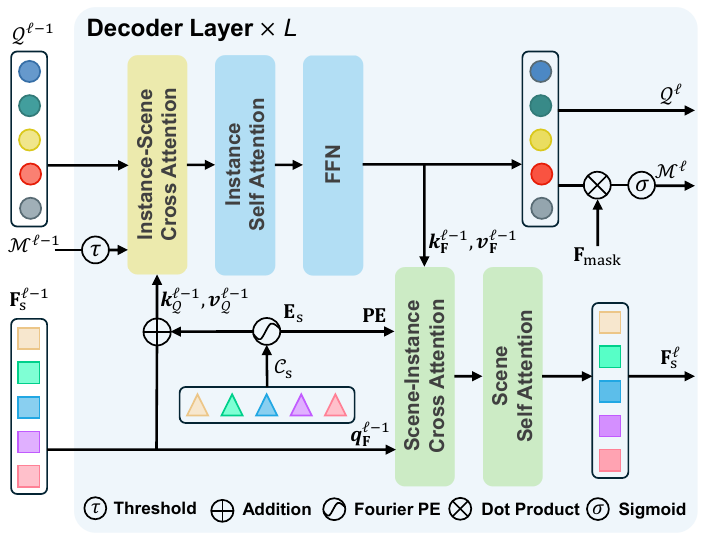}
    \caption{\textbf{Geometric-enhanced Interleaving Transformer (\git) decoder.} The diagram illustrates the detailed structure of our designed decoder. The decoder consists of $L$ layers and employs an alternating update scheme to capture fine-grained features. In each layer, the instance queries $\gQ$, and scene features $\rmF_{\texttt{s}}$ are iteratively refined by incorporating shifted coordinates embedding $\rmE_{\texttt{s}}$. The refined instance queries are then utilized to predict masks $\gM$ and categories $\boldsymbol{p}$.}
    \label{fig:decoder}
\end{figure}

By refining the coordinates, we bring voxels belonging to the same instance closer, promoting the similarity between corresponding features. It is important to note that large-scale scene point clouds typically contain numerous voxels, although they carry rich information, directly leveraging them as scene features for the transformer decoder can be computationally demanding under the quadratic complexity of the attention mechanism. Therefore, we further cluster voxels into superpoints $\gS = \{\gC_{\texttt{s}}, \rmF_{\texttt{s}}\}$ aiming to reduce the complexity:
\begin{equation}
    \gC^{i}_{\texttt{s}} = \frac{1}{|\gN_i|} \sum_{j \in \gN_i} \gG^{-1}(\hat{\gC}^{j}_{\texttt{ref}}),
    \label{eq:superpoint}
\end{equation}
where $\gN_i$ is the set of point indices belonging to the $i$-th superpoint, and $\gG^{-1}(\cdot)$ is the inverse mapping. We apply a similar operation following a non-linear transformation $\phi_{\texttt{s}}$ to features $\rmF$ to obtain $\rmF_{\texttt{s}}$. $\gC_{\texttt{s}} \in \R^{n_{\texttt{s}} \times 3}$ and $\rmF_{\texttt{s}} \in \R^{n_{\texttt{s}} \times d}$ are superpoint coordinates and features, respectively. We consider $\rmF_{\texttt{s}}$ as scene features and send them with the overall instance query set $\gQ$ into our tailored interleaving transformer decoder illustrated in \figref{fig:decoder}.

Our decoder consists of $L$ layers, each comprising a query refinement block (\cf \figref{fig:decoder} upper part) and a scene feature update block (\cf \figref{fig:decoder} lower part). These blocks are updated alternately to enhance geometric information and capture fine-grained details. The query refinement block is specifically designed to update the mix instance query $\gQ$ by progressively attending to the scene features $\rmF_{\texttt{s}}$. Unlike previous transformer-based methods~\cite{sun2023spformer, kolodiazhnyi2023of3d} that overlook the position of superpoints, we incorporate the refined superpoint coordinates to provide geometric information, helping better instance localization. Practically, we exploit Fourier positional encoding~\cite{tancik2020fourier} to get the position embedding $\rmE_{\texttt{s}} \in \R^{n_{\texttt{s}} \times d}$:
\begin{equation}
    \rmE_{\texttt{s}} = \phi_{\texttt{E}}\left(\mathbf{Fourier}(\gC_{\texttt{s}})\right).
    \label{eq:pos_enc}
\end{equation}
We further add resulting embedding to original scene features as geometry reinforced features $\rmF^{\ell-1}_{\texttt{s}} + \rmE_{\texttt{s}}$. Here, compared to static positional information, our $\rmE_{\texttt{s}}$ is derived from the previously predicted bias vector in \eqref{eq:voxel_bias}, which means the positional encoding is dynamically updated based on the estimated bias vectors, allowing for additional information exchange between the backbone and decoder. This technique makes the auxiliary tasks and instance segmentation complementary. By applying linear projection to the enhanced features, we obtain keys $\vk^{\ell-1}_{\gQ}$ and values $\vv^{\ell-1}_{\gQ}$ respectively for the queries $\gQ^{\ell-1}$. We utilize masked cross-attention proposed in~\cite{cheng2022mask2former} to refine the queries constrained by attention mask $\gA^{\ell-1}$ from the last predicted instance mask $\gM^{\ell-1}$:
\begin{equation}
    \gA^{\ell-1}_{i, j} = -\infty \cdot [\gM^{\ell-1}_{i, j} < \tau],
    \label{eq:mask_att}
\end{equation}
where $[\cdot]$ denotes Iverson Brackets and $\tau$ is a threshold. This operation can ensure each query only focuses on relevant context information, improving model robustness. Self-attention and feed-forward networks are also applied to facilitate the interaction between queries, avoiding duplicate instances and capturing discriminative representations. 

Although using superpoint features as global information can reduce computational complexity, we argue that simply pooling voxels into superpoints (\cf \eqref{eq:superpoint}) loses fine-grained features and lacks inter-superpoint communication. Therefore, we introduce a superpoint update block to refine $\rmF^{\ell-1}_{\texttt{s}}$. Benefiting from our semantic-guided mix query initialization scheme, the refined query $\gQ^{\ell}$ already contains fine-grained details from voxel features, which can mitigate the loss of information. By attending to the refined queries $\gQ^{\ell}$ with superpoint position embedding $\rmE_{\texttt{s}}$, we can obtain updated scene features $\rmF^{\ell}_{\texttt{s}}$, which are then passed to the next layer. During the query refinement stage, our decoder iteratively participates in geometric information to emphasize the localization, while during the superpoint update stage, fine-grained features can be captured. Working interleavingly, our proposed decoder can thoroughly interact with instance queries and superpoints to improve performance.

Given refined queries $\gQ^{\ell}$ after each decoder layer and mask-aware features $\rmF_{\texttt{mask}}$, we can get final binary instance masks by:
\begin{equation}
    \gM^{\ell} = \left\{b_{i,j} = \left[\sigma \left( \rmF_{\texttt{mask}} \cdot \phi_{\texttt{m}}(\gQ^{\ell})^\top \right)_{i,j} > \tau \right]\right\}
    \label{eq:mask}
\end{equation}
where $\sigma(\cdot)$ is the sigmoid function, $\tau$ is the same as in \eqref{eq:mask_att}, and $\phi_{\texttt{m}}(\cdot)$ is a normalization layer. Besides, we predict corresponding instance categories $\boldsymbol{p}^{\ell} = \phi_{\texttt{cls}}(\gQ^{\ell})$ implemented by a shared shallow MLP.

\subsection{Loss Function}\label{subsec:loss}
Following exiting transformer-based methods~\cite{sun2023spformer, schult2023mask3d, lu2023query}, we use bipartite graph matching~\cite{karp1990optimal} for instance pairing. Note that we perform matching at every decoder layer. We omit the layer index $\ell$ for simplicity in the following description. The bipartite matching cost between the $i$-th predicted instance and the $j$-th ground truth is defined as:
\begin{equation}
    \begin{split}
        \gU_{i, j} = & -\lambda_{\texttt{cls}} \cdot p_{i,j} + \lambda_{\texttt{bce}} \mathbf{BCE}(\gM_i, \gM^{gt}_j) \\
        & + \lambda_{\texttt{dice}}\mathbf{Dice}(\gM_i, \gM^{gt}_j),
    \end{split}
    \label{eq:cost}
\end{equation}
where $p_{i, j} \subseteq \boldsymbol{p}$ is the predicted probability, $\gM_i$ and $\gM^{gt}_j$ are the $i$-th predicted mask and the $j$-th ground truth mask, respectively. $\lambda_{\texttt{cls}}$, $\lambda_{\texttt{bce}}$, and $\lambda_{\texttt{dice}}$ are hyper-parameters to balance classification, binary cross-entropy $\mathbf{BCE}(\cdot, \cdot)$, and dice $\mathbf{Dice}(\cdot, \cdot)$. We then use Hungarian algorithm~\cite{kuhn1955hungarian} on the cost matrix $\gU$ to find the optimal one-to-one matching between predicted instances and the ground truth instances, formulated as:
\begin{equation}
    \begin{aligned}
    \hat{\boldsymbol{\kappa}}=\arg \min_{\kappa} \sum^{q}_{i=1} \gU_{i, \kappa_{i}} \\
    \text { s.t. } \forall i \neq j, \kappa_{i} \neq \kappa_{j}.
    \end{aligned}
\end{equation}
where $\hat{\boldsymbol{\kappa}}$ is the optimal matching result. Finally, we calculate the classification cross-entropy loss $\gL_{\texttt{cls}}$, binary cross-entropy loss $\gL_{\texttt{mask}}$, and dice loss $\gL_{\texttt{dice}}$ for each matched pair. The overall loss function is given by:
\begin{equation}
    \begin{split}
    \gL = & \lambda_{\texttt{aux}}(\gL_{\texttt{sem}} + \gL_{\texttt{geo}}) \\
    & + \sum_{\ell=1}^{L} \lambda_{\texttt{cls}} \gL^{\ell}_{\texttt{cls}} + \lambda_{\texttt{bce}} \gL^{\ell}_{\texttt{bce}} + \lambda_{\texttt{dice}} \gL^{\ell}_{\texttt{dice}},
    \end{split}
    \label{eq:loss}
\end{equation}
where $\lambda_{\texttt{aux}}$ is used to balance the auxiliary loss and main loss.

\section{Experiment}\label{sec:experiment}

\subsection{Datasets}
To verify the effectiveness of our proposed method, we conduct experiments on three indoor datasets: \scannet~\cite{dai2017scannet}, \scannetp~\cite{rozenberszki2022lground}, and \scannetpp~\cite{yeshwanth2023scannet++}. \textit{\scannet}~\cite{dai2017scannet} is a widely used dataset containing 1201 fully labeled indoor scans for training and 312 scans for validation, respectively. Each scene is carefully annotated with 20 semantic categories, out of which 18 are instance classes (excluding \textit{wall} and \textit{floor}). \scannetp~\cite{rozenberszki2022lground} is an extended version of \scannet with more fine-grained annotations and long-tail distribution covering 198 classes for 3D instance segmentation evaluation. We use the training and validation splits provided by the official toolkit for both datasets. \textit{\scannetpp}~\cite{yeshwanth2023scannet++} is a recently introduced indoor dataset providing sub-millimeter resolution 3D scan geometry along with 84 categories of dense annotations for evaluating instance segmentation quality. The dataset is divided into training, validation, and testing splits of 360, 50, and 50 scenes. Compared to previous datasets, \scannetpp presents more challenges due to its large-scale and complicated layouts. It provides more fine-grained details and high-quality scans, making it a realistic and demanding dataset for evaluating 3D instance segmentation.

% \textit{\sdis}~\cite{armeni2016s3dis} dataset consists of 6 large-scale indoor areas with 271 rooms from three different buildings, and the scans are annotated by 13 instance categories. We follow previous works to assess the segmentation performance on scenes from Area 5.

\subsection{Evaluation Metrics}
For evaluating the performance of our framework, we utilize standard average precision mAP, AP$_{50}$ and AP$_{25}$, which are commonly used in point cloud instance segmentation tasks~\cite{dai2017scannet, rozenberszki2022lground, yeshwanth2023scannet++}. AP$_{50}$ and AP$_{25}$ represent the scores obtained with intersection over union (IoU) thresholds of 50\% and 25\%, respectively, while mAP is calculated as the average score with a set of IoU thresholds from 50\% to 95\% with an increased step size of 5\%. It is noted that higher values of these metrics indicate superior model performance. In addition to performance metrics, we also report the model size and average inference time on the real-world \scannet validation set to evaluate the model efficiency.

\begin{table*}[t]
    \begin{minipage}{0.98\textwidth}
    \centering
        \caption{Comparison with state-of-the-art methods on \scannet~\cite{dai2017scannet} hidden test set. The best results are shown in \textbf{bold}. Results are assessed on July 10th, 2024.} \label{tab:scannet_test}
        \tablestyle{1.5pt}{1.05}
        \begin{tabular}{x{27mm}|x{6mm}x{6mm}|x{6mm}x{6mm}x{6mm}x{6mm}x{6mm}x{6mm}x{6mm}x{6mm}x{6mm}x{6mm}x{6mm}x{6mm}x{6mm}x{6mm}x{6mm}x{6mm}x{6mm}x{6mm}}
    \toprule
    \multirow{3}{*}{Methods} & \multirow{3}{*}{mAP} & \multirow{3}{*}{AP$_{50}$} & \multirow{3}{*}{\rotatebox{90}{bath}} & \multirow{3}{*}{\rotatebox{90}{bed}} & \multirow{3}{*}{\rotatebox{90}{bkshf}} & \multirow{3}{*}{\rotatebox{90}{cabinet}} & \multirow{3}{*}{\rotatebox{90}{chair}} & \multirow{3}{*}{\rotatebox{90}{counter}} & \multirow{3}{*}{\rotatebox{90}{curtain}} & \multirow{3}{*}{\rotatebox{90}{desk}} & \multirow{3}{*}{\rotatebox{90}{door}} & \multirow{3}{*}{\rotatebox{90}{other}} & \multirow{3}{*}{\rotatebox{90}{picture}} & \multirow{3}{*}{\rotatebox{90}{fridge}} & \multirow{3}{*}{\rotatebox{90}{s. cur.}} & \multirow{3}{*}{\rotatebox{90}{sink}} & \multirow{3}{*}{\rotatebox{90}{sofa}} & \multirow{3}{*}{\rotatebox{90}{table}} & \multirow{3}{*}{\rotatebox{90}{toilet}} & \multirow{3}{*}{\rotatebox{90}{window}} \\ 
& \multicolumn{1}{l}{} & \multicolumn{1}{l|}{} & \multicolumn{1}{l}{} & \multicolumn{1}{l}{} & \multicolumn{1}{l}{} & \multicolumn{1}{l}{} & \multicolumn{1}{l}{} & \multicolumn{1}{l}{} & \multicolumn{1}{l}{} & \multicolumn{1}{l}{} & \multicolumn{1}{l}{} & \multicolumn{1}{l}{} & \multicolumn{1}{l}{} & \multicolumn{1}{l}{} & \multicolumn{1}{l}{} & \multicolumn{1}{l}{} & \multicolumn{1}{l}{} & \multicolumn{1}{l}{} & \multicolumn{1}{l}{} \\
& \multicolumn{1}{l}{} & \multicolumn{1}{l|}{} & \multicolumn{1}{l}{} & \multicolumn{1}{l}{} & \multicolumn{1}{l}{} & \multicolumn{1}{l}{} & \multicolumn{1}{l}{} & \multicolumn{1}{l}{} & \multicolumn{1}{l}{} & \multicolumn{1}{l}{} & \multicolumn{1}{l}{} & \multicolumn{1}{l}{} & \multicolumn{1}{l}{} & \multicolumn{1}{l}{} & \multicolumn{1}{l}{} & \multicolumn{1}{l}{} & \multicolumn{1}{l}{} & \multicolumn{1}{l}{} & \multicolumn{1}{l}{} \\
    \midrule
    PointGroup~\cite{jiang2020pointgroup} & 40.7 & 63.6 & 63.9 & 49.6 & 41.5 & 24.3 & 64.5 & 2.1 & 57.0 & 11.4 & 21.1 & 35.9 & 21.7 & 42.8 & 66.6 & 25.6 & 56.2 & 34.1 & 86.0 & 29.1 \\

    HAIS~\cite{chen2021hais} & 45.7 & 69.9 & 70.4 & 56.1 & 45.7 & 36.4 & 67.3 & 4.6 & 54.7 & 19.4 & 30.8 & 42.6 & 28.8 & 45.4 & 71.1 & 26.2 & 56.3 & 43.4 & 88.9 & 34.4 \\

    OccuSeg~\cite{han2020occuseg} & 48.6 & 67.2 & 80.2 & 53.6 & 42.8 & 36.9 & 70.2 & 20.5 & 33.1 & 30.1 & 37.9 & 47.4 & 32.7 & 43.7 & \textbf{86.2} & 48.5 & 60.1 & 39.4 & 84.6 & 27.3 \\

    TD3D~\cite{kolodiazhnyi2023td3d} & 48.9 & 75.1 & 85.2 & 51.1 & 43.4 & 32.2 & 73.5 & 10.1 & 51.2 & 35.5 & 34.9 & 46.8 & 28.3 & 51.4 & 67.6 & 26.8 & 67.1 & 51.0 & 90.8 & 32.9 \\
    
    SoftGroup~\cite{vu2022softgroup} & 50.4 & 76.1 & 66.7 & 57.9 & 37.2 & 38.1 & 69.4 & 7.2 & \textbf{67.7} & 30.3 & 38.7 & 53.1 & 31.9 & 58.2 & 75.4 & 31.8 & 64.3 & 49.2 & 90.7 & 38.8 \\

    SSTNet~\cite{liang2021sstnet} & 50.6 & 69.8 & 73.8 & 54.9 & 49.7 & 31.6 & 69.3 & 17.8 & 37.7 & 19.8 & 33.0 & 46.3 & 57.6 & 51.5 & 85.7 & \textbf{49.4} & 63.7 & 45.7 & 94.3 & 29.0 \\

    \spformer & 54.9 & 77.0 & 74.5 & 64.0 & 48.4 & 39.5 & 73.9 & 31.1 & 56.6 & 33.5 & 46.8 & 49.2 & 55.5 & 47.8 & 74.7 & 43.6 & 71.2 & 54.0 & 89.3 & 34.3 \\

    ISBNet~\cite{ngo2023isbnet} & 55.9 & 75.7 & 93.9 & 65.5 & 38.3 & 42.6 & 76.3 & 18.0 & 53.4 & 38.6 & 49.9 & 50.9 & \textbf{62.1} & 42.7 & 70.4 & 46.7 & 64.9 & 57.1 & 94.8 & \textbf{40.1} \\
    
    Mask3D~\cite{schult2023mask3d} & 56.6 & 78.0 & 92.6 & 59.7 & 40.8 & 42.0 & 73.7 & 23.9 & 59.8 & 38.6 & 45.8 & \textbf{54.9} & 56.8 & \textbf{71.6} & 60.1 & 48.0 & 64.6 & 57.5 & 92.2 & 36.4 \\

    OneFormer3D~\cite{kolodiazhnyi2023of3d} & 56.6 & \textbf{80.1} & 78.1 & 69.7 & \textbf{56.2} & 43.1 & 77.0 & 33.1 & 40.0 & 37.3 & 52.9 & 50.4 & 56.8 & 47.5 & 73.2 & 47.0 & \textbf{76.2} & 55.0 & 87.1 & 37.9 \\

    MAFT~\cite{lai2023mask} & 57.8 & 77.4 & 77.8 & 64.9 & 52.0 & 44.9 & 76.1 & 25.3 & 58.4 & 39.1 & \textbf{53.0} & 47.2 & 61.7 & 49.9 & 79.5 & 47.3 & 74.5 & 54.8 & 96.0 & 37.4 \\

    QueryFormer~\cite{lu2023query} & 58.3 & 78.7 & 92.6 & \textbf{70.2} & 39.3 & \textbf{50.4} & 73.3 & 27.6 & 52.7 & 37.3 & 47.9 & 53.4 & 53.3 & 69.7 & 72.0 & 43.6 & 74.5 & 59.2 & 95.8 & 36.3 \\
    \midrule
    \cellcolor[HTML]{efefef}\ours(Ours) & \cellcolor[HTML]{efefef}\textbf{58.6} & \cellcolor[HTML]{efefef}79.9 & \cellcolor[HTML]{efefef}\textbf{100.0} & \cellcolor[HTML]{efefef}59.3 & \cellcolor[HTML]{efefef}44.0 & \cellcolor[HTML]{efefef}48.0 & \cellcolor[HTML]{efefef}\textbf{77.1} & \cellcolor[HTML]{efefef}\textbf{34.5} & \cellcolor[HTML]{efefef}43.7 & \cellcolor[HTML]{efefef}\textbf{44.4} & \cellcolor[HTML]{efefef}49.5 & \cellcolor[HTML]{efefef}54.8 & \cellcolor[HTML]{efefef}57.9 & \cellcolor[HTML]{efefef}62.1 & \cellcolor[HTML]{efefef}72.0 & \cellcolor[HTML]{efefef}40.9 & \cellcolor[HTML]{efefef}71.2 & \cellcolor[HTML]{efefef}\textbf{59.3} & \cellcolor[HTML]{efefef}\textbf{96.0} & \cellcolor[HTML]{efefef}39.5 \\
    \bottomrule
\end{tabular}
    \end{minipage}
\end{table*}

\subsection{Implementation and Training Details}
We implement our \ours based on the Pointcept~\cite{pointcept2023} toolkit by PyTorch. Following previous work~\cite{kolodiazhnyi2023of3d, shin2023spherical}, we train our model for 510 epochs using the AdamW optimizer with a weight decay of 0.05. We employ a polynomial learning rate scheduler with a base value of 0.9 for the initialized learning rate $3e^{-4}$, but for voxel-wise heads, we set the learning rate to $3e^{-3}$. The model is trained using 4 NVIDIA RTX 4090 GPUs with a batch size of 12. 

For a fair comparison with prior methods, we provide two versions of our model: \ours and \ours-L, specifically designed for \scannet and \scannetp datasets. The only difference between them lies in the choice of feature extractor architecture. \ours adopts a 5-layer Sparse Convolution U-Net~\cite{graham2017submanifold} as the backbone for both datasets, following~\cite{sun2023spformer, lai2023mask, kolodiazhnyi2023of3d}. While \ours-L uses a 7-layer Sparse Convolution U-Net and Res16UNet34C of MinkowskiEngine~\cite{choy2019minkowski} for \scannet and \scannetp, respectively, following~\cite{schult2023mask3d, kolodiazhnyi2023of3d, shin2023spherical, vu2022softgroup, ngo2023isbnet}. But for \scannetpp, we just provide a smaller version. The input point clouds are voxelized with a voxel size of 0.02m for all experiments, and we utilize graph-based over-segmentation~\cite{felzenszwalb2004efficient} to cluster points into superpoints. We apply standard augmentation techniques for point coordinates, including random dropout, horizontal flipping, random rotation around the z-axis, random translation, scaling, and elastic distortion. Our color augmentations include random jittering and auto-contrast following normalization. During the training process, we randomly crop $250k$ points from the original input for each scene in \scannet and \scannetp to reduce the memory consumption while sampling 0.8$\times$ points for \scannetpp. We set the number of scene-aware queries $q_{s}$ and learnable queries $q_{l}$ to 200 and 200, respectively, and use 3 stacked layers to yield the decoder. For the selection ratio $\alpha$, we set it to 0.4. To balance the loss terms, we set $\lambda_{\texttt{cls}}, \lambda_{\texttt{bce}}, \lambda_{\texttt{dice}}$ and $\lambda_{\texttt{aux}}$ to 0.8, 1.0, 1.0 and 0.4, respectively. The same hyperparameters are used for all experiments unless otherwise stated.

\begin{table}[t]
    \begin{minipage}{0.48\textwidth}
    \centering
        \caption{Comparison with state-of-the-art methods on \scannet~\cite{dai2017scannet} validation split. P, G, and T mean proposal-based, group-based, and transformer-based methods, respectively. The best results are shown in \textbf{bold}, and the second best are \underline{underlined}. %%The parameter sizes are obtained from the respective papers or available official codes.%%
        } \label{tab:scannet}
        \tablestyle{1.5pt}{1.05}
        \begin{tabular}{x{27mm}|x{10mm}|x{10mm}x{10mm}x{10mm}}
    \toprule
    Methods & Types & mAP &AP$_{50}$ & AP$_{25}$ \\
    \midrule
    3D-SIS \cite{hou20193dsis} & P & - & 18.7 & 35.7 \\
    GSPN \cite{yi2019gspn} & P & 19.3 & 37.8  &53.4 \\
    TD3D \cite{kolodiazhnyi2023td3d} & P & 47.3 & 71.2 & 81.9 \\
    Spherical Mask \cite{shin2023spherical} & P &\textbf{62.3} &\underline{79.9} &\underline{88.2} \\
    \midrule
    JSNet++ \cite{zhao2022jsnet++} & G & - & 39.2 & 56.8 \\
    PointGroup \cite{jiang2020pointgroup} & G & 34.8 & 56.9 & 71.3  \\
    SSTNet \cite{liang2021sstnet} & G & 49.4 & 64.3 & 74.0  \\
    HAIS \cite{chen2021hais} & G & 43.5 & 64.4 & 74.6   \\
    SoftGroup \cite{vu2022softgroup} & G &45.8 &67.6 &78.9  \\
    PBNet \cite{zhao2023pbnet} & G & 54.3 & 70.5 & 78.9   \\
    \midrule
    Mask3D \cite{schult2023mask3d} & T & 55.2 & 73.7 & 82.9 \\
    \spformer & T & 56.3 & 73.9 & 82.9 \\
    QueryFormer \cite{lu2023query} & T & 56.5 & 74.2 & 83.3  \\
    EipFormer \cite{zhao2023eipformer} & T & 56.9 & 74.6 & - \\
    3IS-ESSS \cite{al20233d} & T &56.1 &75.0 &83.7 \\
    OneFormer3D \cite{kolodiazhnyi2023of3d} & T &59.3 &78.1 &86.4 \\
    \midrule
    \cellcolor[HTML]{efefef}\ours(Ours) &\cellcolor[HTML]{efefef}T &\cellcolor[HTML]{efefef}58.9 &\cellcolor[HTML]{efefef}78.4 &\cellcolor[HTML]{efefef}86.2 \\
    \cellcolor[HTML]{efefef}\ours-L(Ours) &\cellcolor[HTML]{efefef}T &\cellcolor[HTML]{efefef}\underline{61.0} &\cellcolor[HTML]{efefef}\textbf{81.2} &\cellcolor[HTML]{efefef}\textbf{88.9} \\
    \bottomrule
\end{tabular}
    \end{minipage}
\end{table}

\subsection{Comparisons with State-of-the-art Methods}

\begin{table}[t]
    \begin{minipage}{0.48\textwidth}
    \centering
        \caption{Inference time comparison of different methods. We record the average inference time per scene on \scannet~\cite{dai2017scannet} validation set using a single NVIDIA RTX 4090 GPU for all methods. Sp. Ext. refers to superpoint extraction.} \label{tab:inference}
        \tablestyle{1.5pt}{1.05}
        \begin{tabular}{x{27mm}|x{14mm}x{8mm}x{8mm}x{10mm}|x{8mm}}
    \toprule
    \multirow{2}{*}{Methods} & \multirow{2}{*}{Components} & \multirow{2}{*}{Device} & \multirow{2}{*}{\shortstack{Time\\(ms)}} & \multirow{2}{*}{\shortstack{Total\\ (ms)$\downarrow$}} & \multirow{2}{*}{AP$_{50}\uparrow$} \\
    & & & & & \\
    \midrule
    \multirow{2}{*}{\spformer} & Sp. Ext. & CPU & 151.85 & \multirow{2}{*}{\textbf{193.15}} & \multirow{2}{*}{73.9} \\
    & Model & GPU & 41.30 & & \\
    \midrule
    \multirow{2}{*}{OneFormer3D \cite{kolodiazhnyi2023of3d}} & Sp. Ext. & CPU & 151.85 & \multirow{2}{*}{204.98} & \multirow{2}{*}{78.1} \\
    & Model & GPU & 53.13 & & \\
    \midrule
    \multirow{2}{*}{Spherical Mask \cite{shin2023spherical}} & Sp. Ext. & CPU & 151.85 & \multirow{2}{*}{231.95} & \multirow{2}{*}{\underline{79.9}} \\
    & Model & GPU & 80.10 & \\
    \midrule
    \multirow{2}{*}{\ours(Ours)} & Sp. Ext. & CPU & 151.85 & \multirow{2}{*}{\underline{193.99}} & \multirow{2}{*}{78.4} \\
    & Model & GPU & 42.14 & & \\
    \multirow{2}{*}{\ours-L(Ours)} & Sp. Ext. & CPU & 151.85 & \multirow{2}{*}{200.85} & \multirow{2}{*}{\textbf{81.2}} \\
    & Model & GPU & 49.00 & & \\
    \bottomrule
\end{tabular}
    \end{minipage}
\end{table}

We present a comprehensive comparison of our method with state-of-the-art 3D instance segmentation models, including proposal-based (P), group-based (G), and transformer-based (T) methods on \scannet validation and hidden test splits as illustrated in \tabref{tab:scannet_test} and \tabref{tab:scannet}. We report average mAP, AP$_{50}$ for the overall categories and class-wise mAP scores on the test set in \tabref{tab:scannet_test}. \ours obtains the highest average mAP of 58.6\% and achieves the best scores for 6 out of 18 classes. The detailed quantitative results in \tabref{tab:scannet} demonstrate that \ours-L achieves the best performance in terms of AP$_{50}$ and AP$_{25}$, surpassing other methods by a margin of 1.3\% and 0.7\%, respectively. Although our model is slightly inferior to \sph in terms of mAP, it is worth noting that our method significantly improves inference speed by 31ms per scene illustrated in \tabref{tab:inference}, making it well-suited for latency-sensitive scenarios. Our speed advantage is attributed to our model's end-to-end design, whereas \sph relies on a coarse-to-fine strategy involving complex point migration and mask assembly operations. Furthermore, compared with other transformer-based counterparts, our variant \ours achieves equivalent performance as the state-of-the-art OneFormer3D~\cite{kolodiazhnyi2023of3d}, but with fewer parameters and lower latency. This advantage is primarily due to our proposed geometric enhanced interleaving decoder, which reduces the dependence on heavy transformer layers.

We then evaluate our method on \scannetp dataset to verify its robustness and generalization, and the results are showcased in \tabref{tab:scannetp}. Our \ours-L, engaged with \ours, consistently outperforms other methods across all metrics, demonstrating its superiority in handling sophisticated semantics and long-tailed distributions. Concretely, \ours-L achieves 1.1\% and 2.3\% improvements in mAP and AP$_{50}$, respectively. In \tabref{tab:scannetpp}, we benchmark \ours against leading segmentation algorithms on \scannetpp dataset for the effectiveness in processing large-scale and high-fidelity scenes. Our method achieves state-of-the-art performance on both the validation and hidden test sets, achieving an AP$_{50}$ of 37.5\% and 41.1\%, respectively.

\begin{table}[t]
    \begin{minipage}{0.48\textwidth}
    \centering
        \caption{Comparison with state-of-the-art methods on \scannetp~\cite{rozenberszki2022lground} validation set. The best results are shown in \textbf{bold}, and the second best are \underline{underlined}.} \label{tab:scannetp}
        \tablestyle{1.5pt}{1.05}
        \begin{tabular}{x{27mm}|x{10mm}x{10mm}x{10mm}}
    \toprule
    Methods & mAP &AP$_{50}$ & AP$_{25}$ \\
    \midrule
    PointGroup \cite{jiang2020pointgroup} & - & 24.5 & - \\
    SPFormer \cite{sun2023spformer} & 25.2 & 33.8 & 39.6 \\
    TD3D \cite{kolodiazhnyi2023td3d} & 23.1 & 34.8 & 40.4 \\
    Mask3D \cite{schult2023mask3d} & 27.4 & 37.0 & 42.3 \\
    QueryFormer \cite{lu2023query} & 28.1 & 37.1 & 43.4 \\
    \midrule
    \cellcolor[HTML]{efefef}\ours(Ours) &\cellcolor[HTML]{efefef}\underline{28.9} &\cellcolor[HTML]{efefef}\underline{38.6} &\cellcolor[HTML]{efefef}\underline{43.6} \\
    \cellcolor[HTML]{efefef}\ours-L(Ours) &\cellcolor[HTML]{efefef}\textbf{29.2} &\cellcolor[HTML]{efefef}\textbf{39.4} &\cellcolor[HTML]{efefef}\textbf{44.2} \\
    \bottomrule
\end{tabular}
    \end{minipage}
\end{table}

\begin{table}[t]
    \begin{minipage}{0.48\textwidth}
    \centering
        \caption{Comparison with state-of-the-art methods on \scannetpp~\cite{yeshwanth2023scannet++} benchmark. The best results are shown in \textbf{bold}, and the second best results are \underline{underlined}. Results on the hidden test set are assessed on June 24th, 2024. $^\dagger$ denotes metrics are reported by~\cite{yeshwanth2023scannet++}.} \label{tab:scannetpp}
        \tablestyle{1.5pt}{1.05}
        \begin{tabular}{x{27mm}|x{8mm}x{8mm}x{8mm}|x{8mm}x{8mm}x{8mm}}
    \toprule
    \multicolumn{1}{c}{\textit{Instance Seg.}} & \multicolumn{3}{c}{\textbf{\scannetpp \textit{Val}}} & \multicolumn{3}{c}{\textbf{\scannetpp \textit{Test}}} \\
    \cmidrule(lr){1-1} \cmidrule(lr){2-4} \cmidrule(lr){2-4} \cmidrule(lr){5-7}
    Methods & mAP &AP$_{50}$ & AP$_{25}$ & mAP &AP$_{50}$ & AP$_{25}$ \\
    \midrule
    PointGroup$^\dagger$ \cite{jiang2020pointgroup} & - & 14.8 & - & 8.9 & 14.6 & 21.0\\
    HAIS$^\dagger$ \cite{chen2021hais} & - & 16.7 & - & 12.1 & 19.9 & 29.5 \\
    SoftGroup$^\dagger$ \cite{vu2022softgroup} & - & \underline{23.7} & - & 16.7 & 29.7 & 38.9 \\
    BFL \cite{lu2024beyond} & - & - & - & \underline{22.2} & \underline{32.8} & \underline{42.5} \\
    \midrule
    \cellcolor[HTML]{efefef}\ours(Ours) & \cellcolor[HTML]{efefef}\textbf{23.9} & \cellcolor[HTML]{efefef}\textbf{37.5} & \cellcolor[HTML]{efefef}\textbf{46.6} & \cellcolor[HTML]{efefef}\textbf{27.3} & \cellcolor[HTML]{efefef}\textbf{41.0} & \cellcolor[HTML]{efefef}\textbf{48.4} \\
    \bottomrule
\end{tabular}
    \end{minipage}
\end{table}

\begin{figure*}[!htbp] 
    \centering
    \centering
    \includegraphics[width=0.98\textwidth]{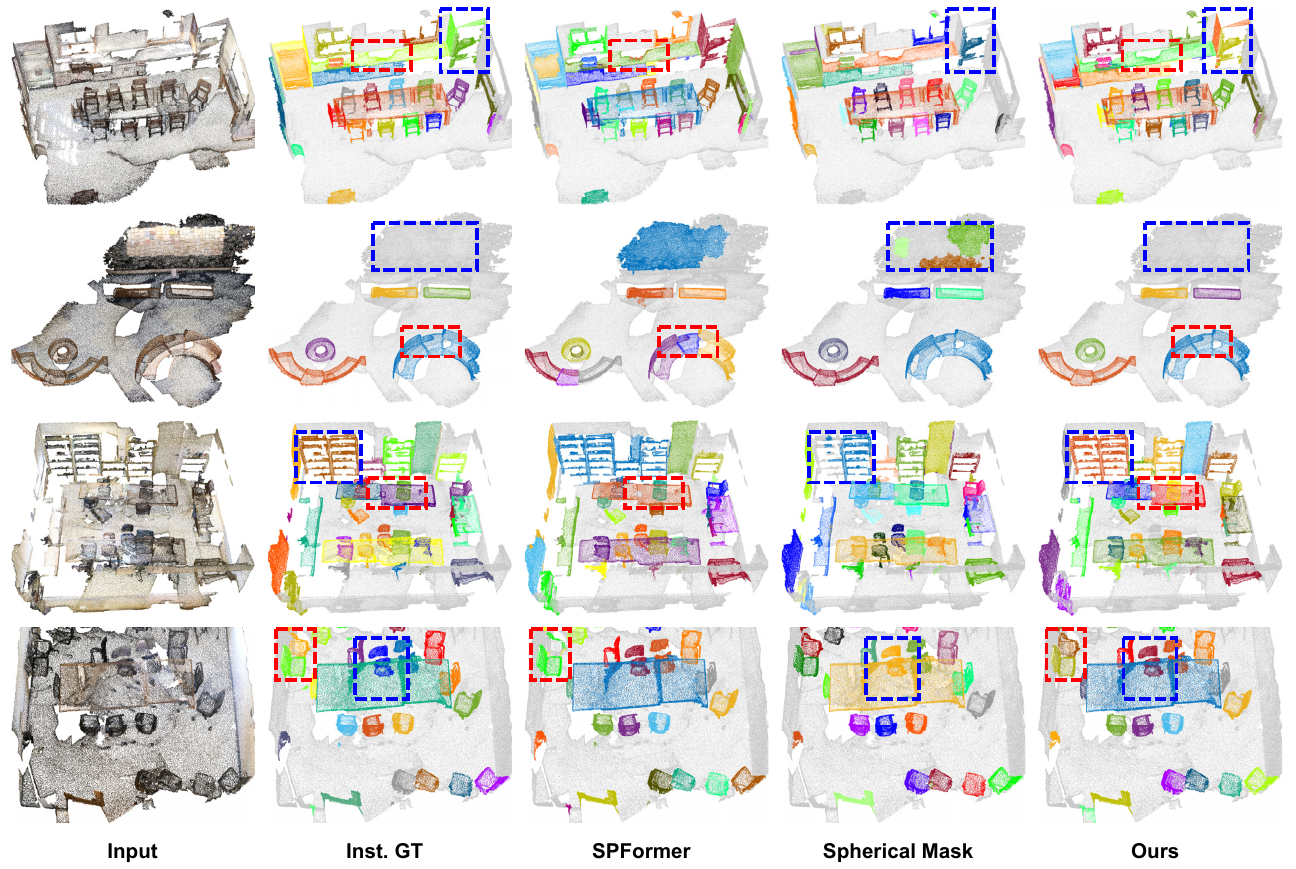}
    \caption{\textbf{Visualization comparison on \scannet validation split.} We visualize the instance segmentation results of \ours (ours), \spformer, and \sph. Inst. GT means instance ground truth, and different colors indicate different instance IDs. The comparison with \spformer is highlighted in \textcolor{red}{red}, while the comparison with \sph is highlighted in \textcolor{blue}{blue}.}
    \label{fig:viz_scannet}
\end{figure*}

\subsection{Ablation Studies}
We conduct comprehensive ablation studies of \ours on \scannet validation set, focusing on evaluating the impact of core designs within our framework. In \tabref{tab:ablation_query}, we explore various combinations of query initialization and decoder designs. Considering alternative parametric learnable query, FPS-based non-parametric query, vanilla transformer decoder, as well as our proposed semantic-guided mix query (\smq) initialization and geometric-enhanced interleaving transformer (\git) decoder, we assess 7 different variants besides our method. When combining with the vanilla transformer decoder, we observe that our proposed \smq (\#7) achieves a 0.5\% improvement in terms of AP$_{50}$ compared to learnable query (\#1) and FPS-based query (\#2). Since our \smq is designed in a hybrid form, we assemble the FPS-based query with the learnable query while maintaining the same number of queries to investigate \smq's impact. Results of \#5 and \#7 show that our novel paradigm slightly improves mAP, indicating that our novel paradigm assists the ability to capture more informative context. Moreover, we validate the effectiveness of \git by incorporating it with different query initialization methods in variants \#2, \#4, and \#6. The results clearly show that our interleaving mechanism significantly boosts the performance by 1.0\% and 0.5\% in mAP and AP$_{50}$, confirming the superior capability of our model in enhancing instance localization.

\begin{table}[t]
    \begin{minipage}{0.48\textwidth}
    \centering
        \caption{Variants of query initialization and decoder designs. Learn., FPS, \smq, and \git indicate learnable query, FPS-based query, semantic guided mix query, and geometric enhanced interleaving decoder, respectively. Variants not chosen \git are trained with the vanilla transformer decoder. The best results are shown in \textbf{bold}, and the second best results are \underline{underlined}.} \label{tab:ablation_query}
        \tablestyle{1.5pt}{1.05}
        % \begin{tabular}{x{8mm}|x{8mm}x{8mm}x{8mm}x{11mm}|x{10mm}x{10mm}x{10mm}}
%     \toprule
%     \# & Learn. & FPS & \smq & w/ \git & mAP &AP$_{50}$ & AP$_{25}$ \\
%     \midrule
%     1 & \YesV & \NoX & \NoX & \NoX & 56.6 & 76.6 & 85.2 \\
%     2 & \NoX & \YesV & \NoX & \NoX & 57.2 & 77.1 & 85.7 \\
%     3 & \YesV & \NoX & \NoX & \YesV & 57.8 & 77.3 & 85.7 \\
%     4 & \NoX & \YesV & \NoX & \YesV & 56.5 & 77.8 & 85.8 \\
%     5 & \YesV & \YesV & \NoX & \NoX & 57.1 & 77.7 & 85.6 \\
%     6 & \YesV & \YesV & \NoX & \YesV & \underline{57.9} & \underline{77.9} & \underline{86.1} \\
%     7 & \NoX & \NoX & \YesV & \NoX & 57.9 & 77.6 & 85.6 \\
%     \midrule
%     \cellcolor[HTML]{efefef}Ours & \cellcolor[HTML]{efefef}\NoX & \cellcolor[HTML]{efefef}\NoX & \cellcolor[HTML]{efefef}\YesV & \cellcolor[HTML]{efefef}\YesV & \cellcolor[HTML]{efefef}\textbf{58.9} & \cellcolor[HTML]{efefef}\textbf{78.4} & \cellcolor[HTML]{efefef}\textbf{86.2} \\
%     \bottomrule
% \end{tabular}

\begin{tabular}{x{8mm}|x{8mm}x{8mm}x{8mm}x{11mm}|x{10mm}x{10mm}x{10mm}}
    \toprule
    \# & Learn. & FPS & \smq & w/ \git & mAP &AP$_{50}$ & AP$_{25}$ \\
    \midrule
    1 & \YesV &  &  &  & 56.6 & 76.6 & 85.2 \\
    2 &  & \YesV &  &  & 57.2 & 77.1 & 85.7 \\
    3 & \YesV &  &  & \YesV & 57.8 & 77.3 & 85.7 \\
    4 &  & \YesV &  & \YesV & 56.5 & 77.8 & 85.8 \\
    5 & \YesV & \YesV &  &  & 57.1 & 77.7 & 85.6 \\
    6 & \YesV & \YesV &  & \YesV & \underline{57.9} & \underline{77.9} & \underline{86.1} \\
    7 &  &  & \YesV &  & 57.9 & 77.6 & 85.6 \\
    \midrule
    \cellcolor[HTML]{efefef}Ours & \cellcolor[HTML]{efefef} & \cellcolor[HTML]{efefef} & \cellcolor[HTML]{efefef}\YesV & \cellcolor[HTML]{efefef}\YesV & \cellcolor[HTML]{efefef}\textbf{58.9} & \cellcolor[HTML]{efefef}\textbf{78.4} & \cellcolor[HTML]{efefef}\textbf{86.2} \\
    \bottomrule
\end{tabular}
    \end{minipage}
\end{table}

As depicted in \tabref{tab:ablation_detail}, we perform a series of experiments by subsequently removing different components, including geometric enhancement (w/ Geo.), scene-aware queries $\rmQ^{s}$, and learnable queries $\rmQ^{l}$. This step-by-step procedure allows us to assess the individual contributions of each component to the overall results. Upon discarding geometric enhancement, we observe a noticeable performance drop across all metrics, particularly in terms of mAP, indicating the necessity of our proposed progressive geometric refinement mechanism. In addition, the table shows that our implicitly initialized scene-aware queries $\rmQ^{s}$ play a crucial role in improving the accuracy of instance segmentation, while the removal of learnable queries $\rmQ^{l}$ has a relatively minor impact. This suggests that our scene-aware queries are more effective in capturing instance-level information, and a combination of both queries can further enhance the capability of our interleaving decoder for better aggregating instance features from the global context. As discussed in \secref{subsec:decoder} (\cf \eqref{eq:voxel_bias}), we argue that refining the coordinates by shifting learned bias can intuitively lead to more discriminative voxel representations, ultimately resulting in improved 3D instance segmentation. To validate this claim, we conduct an ablation study by replacing the bias estimation with raw coordinate regression, and the quantitative results are presented in \tabref{tab:ablation_coord}. It is evident that our design can significantly improve the performance by 1.5\% for mAP.

\begin{table}[t]
    \begin{minipage}{0.48\textwidth}
    \centering
        \caption{Effect of removing each component in \ours. Geo. indicates progressively geometric enhancement. The best results are shown in \textbf{bold}.} \label{tab:ablation_detail}
        \tablestyle{1.5pt}{1.05}
        \begin{tabular}{x{10mm}x{10mm}x{10mm}|x{10mm}x{10mm}x{10mm}}
    \toprule
    $\rmQ^{s}$ & $\rmQ^{l}$ & w/ Geo. & mAP &AP$_{50}$ & AP$_{25}$ \\
    \midrule
    \cellcolor[HTML]{efefef}\YesV & \cellcolor[HTML]{efefef}\YesV & \cellcolor[HTML]{efefef}\YesV & \cellcolor[HTML]{efefef}\textbf{58.9} & \cellcolor[HTML]{efefef}\textbf{78.4} & \cellcolor[HTML]{efefef}\textbf{86.2} \\
    \YesV & \YesV & \NoX & 57.6 & 78.2 & 86.1 \\   
    \YesV & \NoX & \YesV & 57.8 & 77.9 & 86.1 \\
    \NoX & \YesV  & \YesV & 56.6 & 77.8 & 85.6 \\ 
    \bottomrule
\end{tabular}
    \end{minipage}
\end{table}

\begin{table}[t]
    \begin{minipage}{0.48\textwidth}
    \centering
        \caption{Comparison of different geometric estimation methods.} \label{tab:ablation_coord}
        \tablestyle{1.5pt}{1.05}
        \begin{tabular}{x{30mm}|x{10mm}x{10mm}x{10mm}}
    \toprule
    Methods & mAP & AP$_{50}$ & AP$_{25}$\\
    \midrule
    Coordinate Regression & 57.4 & 77.4 & 85.0 \\
    \cellcolor[HTML]{efefef}Bias Estimation & \cellcolor[HTML]{efefef}\textbf{58.9} & \cellcolor[HTML]{efefef}\textbf{78.4} & \cellcolor[HTML]{efefef}\textbf{86.2} \\
    \textit{Improvement} & \textit{+1.5} & \textit{+1.0} & \textit{+1.2} \\
    \bottomrule
\end{tabular}
    \end{minipage}
\end{table}

\begin{figure*}[!htbp] 
    \centering
    \centering
    \includegraphics[width=0.98\textwidth]{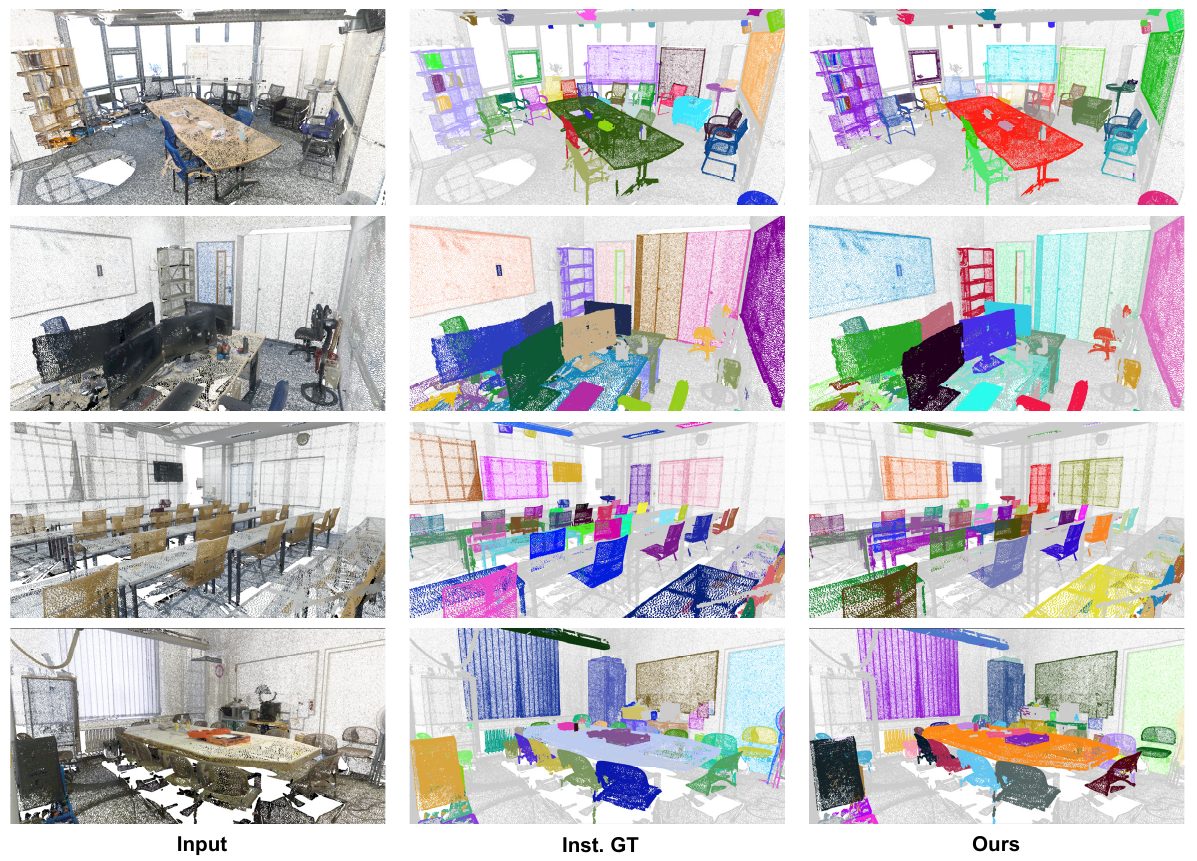}
    \caption{\textbf{Qualitative results of \scannetpp validation set.} We present 4 representative examples selected from \scannetpp validation set to showcase the input point clouds, instance ground truth, and the segmentation results of \ours. The visualization comprehensively illustrates our method's capability in handling large-scale and high-fidelity scenes.}
    \label{fig:viz_scannetpp}
\end{figure*}

We then analyze the influence of selection ratio $\alpha$ defined in Alg. \ref{alg:query}. We vary $\alpha$ from 0.2 to 1.0 with a step size of 0.2 
to explore its impact, where a value of 1.0 means the utilization of all voxels for scene-aware queries. The results presented in \tabref{tab:ablation_alpha} demonstrate that the model achieves the best performance when $\alpha$ is set to 0.4, which indicates that a moderate selection ratio can effectively filter out disruptive and redundant voxels to balance the trade-off between the global context and local details. In \tabref{tab:ablation_params}, we investigate the effect of query numbers $q$ and stacked layers $L$ in the decoder. During these experiments, we maintain the same number of scene-aware and learnable queries. We can observe from the results that the model achieves the worst performance with only one decoder layer, and increasing the query number will not bring significant improvement. While slightly increasing the number of stacked layers can boost the performance, the model with 3 layers and 400 queries achieves the best results. However, further increasing the layers to 6 even degrades the performance. This decline can be attributed to the excessive number of layers that will devastate our interleaving mechanism and introduce more noise to the instance features.

\begin{table}[t]
    \begin{minipage}{0.48\textwidth}
    \centering
        \caption{Impact analysis of selection ratio $\alpha$ on the instance segmentation performance.} \label{tab:ablation_alpha}
        \tablestyle{1.5pt}{1.05}
        \begin{tabular}{x{12mm}|x{12mm}x{12mm}x{12mm}x{12mm}x{12mm}}
    \toprule
    $\alpha$ & 0.2 & \cellcolor[HTML]{efefef}0.4 & 0.6 & 0.8 & 1.0 \\
    \midrule
    mAP & 57.4 & \cellcolor[HTML]{efefef}\textbf{58.9}$^{\textcolor{red}{\scriptscriptstyle\uparrow}\text{\textcolor{red}{1.2}}}$ & \underline{57.7} & 57.0 & 57.0 \\
    AP$_{50}$ & 78.0 & \cellcolor[HTML]{efefef}\textbf{78.4}$^{\textcolor{red}{\scriptscriptstyle\uparrow}\text{\textcolor{red}{0.3}}}$ & \underline{78.1} & 77.9 & 77.9  \\
    AP$_{25}$ & 86.0 & \cellcolor[HTML]{efefef}\underline{86.2}$^{\textcolor{darkgreen}{\scriptscriptstyle\downarrow}\text{\textcolor{darkgreen}{0.7}}}$ & 85.8 & \underline{86.2} & \textbf{86.9}  \\
    \bottomrule
\end{tabular}
    \end{minipage}
\end{table}

\begin{table}[t]
    \begin{minipage}{0.48\textwidth}
    \centering
        \caption{Effect of query numbers and stacked layers in the decoder.} \label{tab:ablation_params}
        \tablestyle{1.5pt}{1.05}
        % \begin{tabular}{x{7mm}|x{7mm}x{7mm}x{7mm}|x{7mm}x{7mm}x{7mm}|x{7mm}x{7mm}x{7mm}}
%     \toprule
%     \textit{Query} &\multicolumn{3}{c}{\textbf{1 Layer}} &\multicolumn{3}{c}{\textbf{3 Layers}} &\multicolumn{3}{c}{\textbf{6 Layers}}\\
%     \cmidrule(lr){1-1}\cmidrule(lr){2-4} \cmidrule(lr){5-7} \cmidrule(lr){8-10}
%     \#num & mAP &AP$_{50}$ & AP$_{25}$ & mAP &AP$_{50}$ & AP$_{25}$ & mAP &AP$_{50}$ & AP$_{25}$\\
%     \midrule
%     200 & 54.7 & 75.4 & 84.4 & \underline{57.6} & 77.4 & \textbf{86.3} & 56.5 & 77.4 & 85.8 \\
%     \midrule
%     400 & 55.8 & 75.6 & 84.4 & \cellcolor[HTML]{efefef}\textbf{58.9} & \cellcolor[HTML]{efefef}\textbf{78.4} & \cellcolor[HTML]{efefef}\underline{86.2} & \underline{57.6} & 77.7 & 85.5 \\
%     \midrule
%     600 & 55.9 & 75.6 & 84.2 & 57.1 & \underline{77.8} & 85.7 & 57.4 & 77.5 & 85.2 \\
%     \bottomrule
% \end{tabular}

\begin{tabular}{x{13mm}|x{7mm}x{7mm}x{7mm}|x{7mm}x{7mm}x{7mm}|x{7mm}x{7mm}x{7mm}}
    \toprule
    \textit{Layers} &\multicolumn{3}{c}{\textbf{1 Layer}} &\multicolumn{3}{c}{\textbf{3 Layers}} &\multicolumn{3}{c}{\textbf{6 Layers}}\\
    \cmidrule(lr){1-1}\cmidrule(lr){2-4} \cmidrule(lr){5-7} \cmidrule(lr){8-10}
    \#num & 200 & 400 & 600 & 200 & \cellcolor[HTML]{efefef}400 & 600 & 200 & 400 & 600 \\
    \midrule
    mAP & 54.7 & 55.8 & 55.9 & \underline{57.6} & \cellcolor[HTML]{efefef}\textbf{58.9} & 57.1 & 56.5 & 57.6 & 57.4 \\
    \midrule
    AP$_{50}$ & 75.4 & 75.6 & 75.6 & 77.4 & \cellcolor[HTML]{efefef}\textbf{78.4} & 77.8 & 77.4 & \underline{77.7} & 77.5 \\
    \midrule
    AP$_{25}$ & 84.4 & 84.4 & 84.2 & \textbf{86.3} & \cellcolor[HTML]{efefef}\underline{86.2} & 85.7 & 85.8 & 85.5 & 85.2 \\
    \midrule
    Params(M) & 12.75 & 12.80 & 12.85 & 15.91 & \cellcolor[HTML]{efefef}15.96 & 16.01 & 20.66 & 20.71 & 20.76 \\
    \bottomrule
\end{tabular}
    \end{minipage}
\end{table}

\subsection{Qualitative Results}
\figref{fig:viz_scannet} illustrates several representative qualitative examples of our method on \scannet validation set, comparing with \spformer and \sph. Empowered by semantic guided and geometric enhanced properties, our method can accurately recognize large convex and complex shapes (row 1 and row 2), such as counter, sofa, and background. For challenging scenes with cluttered and nearby instances (row 3 and row 4), \ours can effectively separate instances with fine-grained details (\eg tables, chairs, and bookshelf). In contrast, \spformer and \sph tend to miss small parts or merge different instances into a single object. Furthermore, to demonstrate the ability of our method in handling large-scale and high-fidelity scenes, we visualize the results on \scannetpp validation set in \figref{fig:viz_scannetpp}. Notably, these examples show that our method can accurately distinguish instances with similar shapes and textures under complex layouts, which is crucial for real-world applications.

\section{Conclusion}\label{sec:conclusion}
This paper presents a novel transformer-based 3D point cloud instance segmentation method namely \ours. The proposed semantic-guided mix query initialization scheme combines implicitly generated scene-aware query from the original input with the learnable query, which overcomes the query initialization dilemma in large-scale 3D scenes. This hybrid strategy cannot only filter out irrelevant information but also empower the model to handle semantically sophisticated scenes. Considering the vanilla transformer decoder struggles to capture fine-grained instance details and relies on heavily stacked layers, we introduce a geometric-enhanced interleaving transformer decoder to update instance queries and global features alternately, progressively incorporating coordinate information for better instance localization. \ours achieves state-of-the-art performance on \scannet and \scannetp datasets, and the latest high-quality \scannetpp instance segmentation benchmark. Comprehensive ablation studies demonstrate the effectiveness of each component in our architecture.

\section*{Acknowledgment}
The research work was conducted in the JC STEM Lab of Machine Learning and Computer Vision funded by The Hong Kong Jockey Club Charities Trust. 

\bibliographystyle{IEEEtran}
\bibliography{IEEEabrv, references}

\end{document}